\documentclass[10pt,twocolumn,letterpaper]{article}

\pdfoutput=1

\usepackage{cvpr}
\usepackage{times}
\usepackage{epsfig}
\usepackage{graphicx}
\usepackage{amsmath}
\usepackage{amssymb}

\usepackage[pagebackref=true,breaklinks=true,letterpaper=true,colorlinks,bookmarks=false]{hyperref}

\cvprfinalcopy 

\ifcvprfinal\fi

\title{Learning Random Kernel Approximations for Object Recognition}

\author{
\begin{tabular}{ccc}
Eduard Gabriel B\u{a}z\u{a}van $^{1}$ & Fuxin Li$^{2}$ & Cristian Sminchisescu$^{2,1}$\\
{\small eduard.bazavan@imar.ro} & {\small fli@cc.gatech.edu} & {\small cristian.sminchisescu@ins.uni-bonn.de}\\
\end{tabular}\\
\\
$^1$Institute of Mathematics of the Romanian Academy\\
$^2$Faculty of Mathematics and Natural Science, University of Bonn\\
$^3$Georgia Institute of Technology\\
}

\begin{document}


\maketitle

\begin{abstract}
Approximations based on random Fourier features have recently emerged as an efficient and formally consistent methodology to design large-scale kernel machines \cite{Rahimi07randomfeatures}. By expressing the kernel as a Fourier expansion, features are generated based on a finite set of random basis projections, sampled from the Fourier transform of the kernel, with inner products that are Monte Carlo approximations of the original kernel. Based on the observation that different kernel-induced Fourier sampling distributions correspond to different kernel parameters, we show that an optimization process in the Fourier domain can be used to identify the different frequency bands that are useful for prediction on training data. Moreover, the application of group Lasso \cite{Yuan06modelselection} to random feature vectors corresponding to a linear combination of multiple kernels, leads to efficient and scalable reformulations of the standard multiple kernel learning model \cite{Varma09}. In this paper we develop the linear Fourier approximation methodology for both single and multiple gradient-based kernel learning and show that it produces fast and accurate predictors on a complex dataset such as the Visual Object Challenge 2011 (VOC2011).
\end{abstract}

\section{Introduction}
The proper choice of kernel function and its hyperparameters are crucial to the success of applying kernel methods to practical applications. These selections span a number of different problems: from choosing a single width parameter in radial basis kernels, scaling different feature dimensions with different weights \cite{Chapelle2002}, to learning a linear or a non-linear combination of multiple kernels (MKL) \cite{Lanckriet2004}. In complicated practical problems such as computer vision, sometimes the need of multiple kernels arises naturally \cite{Vedaldi2009,Kapoor2010}. Images can be represented using descriptors based on shape, color and texture, and these descriptors have different roles in classifying different categories.
In such situations it is in principle easy to design a kernel classifier where each kernel represents one of the descriptors and the classifier would be based on a weighted combination of kernels with category dependent learnt weights.

A natural difficulty in kernel learning is scalability. Kernel methods scale with an already mediocre time complexity of at least $O(N^{2.3})$ with respect to the size $N$ of the training set, but combining multiple kernels or learning the hyperparameters of a single kernel significantly slows down training. Some speed-ups apply for specific kernels, but only in limited scenarios \cite{Maji2008}. In consequence, most of the kernel learning approaches so far are only capable to handle a few thousand training examples at most. This is insufficient for the current age of massive datasets, such as a 11 million images ImageNet, or the 19 million articles within Wikipedia.

An emerging technique that can in principle speed up the costly kernel method while at the same time preserving its non-linear predictive power is the random Fourier feature methodology (RFF) \cite{Rahimi07randomfeatures,Vedaldi2010,li_dagm10}. By sampling components from the frequency space of the kernel using Monte Carlo methods, RFF obtains a bounded, approximate representation of a kernel embedding that may initially span an infinite-dimensional space. Many operations are simplified once such representation is available, the most notable being that any kernel learning algorithm would now scale as $O(N)$, where $N$ is the number of examples. This opens the path for applying kernel methods to the realm of massive datasets. 

In the seminal work on random Fourier features \cite{Rahimi07randomfeatures}, the methodology was developped primarily for radial basis kernels. Recent work \cite{Vedaldi2010,li_dagm10} focused on extending this technique to a number of other useful kernels defined on histogram features (empirical estimates of multinomial distributions), such as the $\chi^2$ and histogram intersection measures. However, the potential of the linear random Fourier methodology for kernel learning remains largely unexplored. In this paper we develop the methodology for learning both single kernel and for multiple kernel combinations in the Fourier domain and show that these produce accurate and efficient models. 
We conduct experiments in visual object recognition, using the difficult PASCAL Visual Object Challenges 2011 dataset, in order to demonstrate the performance of the proposed Fourier kernel learning methodology and compare against non-linear learning algorithms, designed to operate in the original kernel space.

\section{Related work}

Approaches to kernel learning can be broadly classified into methods that estimate the hyper-parameters of a single kernel\cite{GaussianProcessesBook,Chapelle2002,Keerthi2007} and methods that learn the weights of a linear combinations of kernels and possibly their hyper-parameters -- the so-called multiple kernel learning framework or MKL\cite{Szafranski2008,Bach2009,Kloft2009,Vishwanathan2010}.

A popular approach for single kernel learning is the gradient-based method pursued by Chapelle \etal  \cite{Chapelle2002,Keerthi2007}. Keerthi \etal \cite{Keerthi2007} give an efficient algorithm that alternates between learning an SVM and optimizing the feature weights. Cortes \etal \cite{Cortes2010} propose a two-stage method based on a modification of a classical kernel alignment \cite{Cristianini2002} metric and prove a number of learning bounds. Approaches to single kernel learning under a kernel prior have been pursued both as semi-definite programs\cite{Lanckriet2004} and within unconstrained optimization formulations\cite{li_aistats_09}, although in both cases the optimization is involved and complexity is an issue. More recent methods attempt to track the entire regularization path in the non-linear case \cite{Rosset2005,lis_aistats10}.

Multiple kernel learning provides a powerful conceptual framework for both model combination and model selection and has attracted significant research recently. Initial approaches originating with work by Lanckriet \etal \cite{Lanckriet2004} estimated a linear combination of kernels using semi-definite programming. This was reformulated by Bach \etal \cite{Bach2004} as block-norm regularization, reducing the optimization problem to a second order cone program applicable to medium scale problems. More recent methods \cite{Cortes2009} learn a polynomial combination of kernels. A number of approaches have pursued different $l_p$-norms for kernel selection \cite{Szafranski2008,Kloft2009,Vishwanathan2010}. Hierarchical kernel learning approaches like (HKL) \cite{Bach2009}  perform feature selection combinatorially, by choosing kernel combinations obtained by mapping to a directed acyclic graph. The search for such combinations can then be performed in polynomial time.  

A difficulty with many single or multiple kernel learning formulations is their relatively unfavorable scaling properties. When kernels are used, such methods usually scale at least quadratically with the number of examples. Sometimes scaling with the number of kernel parameters and the number of kernels can also become an issue. A formulation of kernel learning within a linear Fourier framework, as pursued in this paper, carries the promise of better scalability and wider applicability to large datasets, while at the same time preserving the non-linear predictive power that makes kernel methods attractive.

\section{Learning Kernels Parameters}

\subsection{Random Fourier Approximations}
Kernels offer substantial power and flexibility to predictive models. Central to their use is the `kernel trick' which allows the design of algorithms that depend only on the inner product of their arguments. This is based on the property of positive definite kernel functions $k(\cdot,\cdot)$ to define a dot product and a lifting $\phi$ so that the dot product between lifted data points can be efficiently computed as $\phi(\mathbf{x})^{\top}\phi(\mathbf{y})=k(\mathbf{x},\mathbf{y})$. Algorithms become linear in a complex feature space induced by $\phi$, but access the data only through evaluations of the kernel $k$ in input space. This makes possible to handle complex or infinite dimensional feature space mappings $\phi$, and indeed these usually give the best results in visual recognition datasets \cite{Gehler2009,lis_cvpr10}. However the very advantage that made kernel methods popular--the kernel trick, requires the manipulation of matrices of pairwise examples. For large datasets, the scaling of kernel methods is at least quadratic in the number of examples. This makes their direct usage impractical beyond datasets of $10^5$ elements. The methodology we pursue is to approximate the kernel function linearly using random feature maps.

Key to the random Fourier methodology is Bochner's theorem, that connects positive definite kernels and their Fourier transforms \cite{RudinFourierGroups,Rahimi07randomfeatures,li_dagm10}. For positive definite translation-invariant kernels $k(\mathbf{x},\mathbf{y}) = k(\mathbf{x}-\mathbf{y})$ on $\mathbb{R}^m$, Bochner's theorem guarantees that every kernel is the inverse Fourier transform of a proper probability distribution $\mu$. Defining $\zeta(\mathbf{x})=e^{j\mathbf{x}^{\top}\boldsymbol{\gamma}}$ (with $j$ the imaginary unit), the following equality holds:

\begin{eqnarray*}
\label{eqn:approximate_kernel}
k_{\boldsymbol{\sigma}}(\mathbf{x},\mathbf{y}) &=& \int_{\mathbb{R}^m} e^{j(\mathbf{x}-\mathbf{y})^{\top}\boldsymbol{\gamma}} d \mu_k(\boldsymbol{\gamma}) \\ &=& \mathbf{E}_\mu[\zeta_{\boldsymbol{\gamma}}(\mathbf{x})\zeta_{\boldsymbol{\gamma}}(\mathbf{y})^*]
\\ &\approx&
\boldsymbol{\phi}_{\boldsymbol{\Gamma}}(\mathbf{x})^{\top}\boldsymbol{\phi}_{\boldsymbol{\Gamma}}(\mathbf{y})
\end{eqnarray*}

where $*$ is the (complex) conjugate, and \begin{equation}\boldsymbol{\phi}_{\Gamma}(\mathbf{x}) = \sqrt{\frac{2}{d}} \left[ \cos \left( \mathbf{x}^{\top} \boldsymbol{\gamma}_{i} + 2\pi b_{i} \right)\right]_{i=1,d} \label{general_feat_formula}\end{equation} is the random feature map at frequencies $\Gamma= \{\boldsymbol{\gamma}_1, \ldots, \boldsymbol{\gamma}_d\}$, with $b \sim \mathrm{U}[0, 2 \pi]$, the uniform distribution in the interval $[0,2\pi]$. We approximate the expectation $\mathbf{E}_\mu[\zeta_{\boldsymbol{\gamma}}(\mathbf{x}) \zeta_{\boldsymbol{\gamma}}(\mathbf{y})^*]$ by means of a Monte-Carlo sample drawn from the distribution $\mu_k$. This is an operation on linear functions with explicit features. The algorithm for the change of representation has two steps: i) Generate $d$ random samples $\Gamma $ from the distribution $\mu_k$; ii) Compute the random projection $\boldsymbol{\phi}_{\Gamma}$ using (\ref{general_feat_formula}), for all training examples. The approximation has the convergence rate of Monte Carlo methods, $O(d^{-1/2})$, dependent on the random sample size, but independent of the input dimension. One usually needs up to a few thousand dimensions to approximate the original kernel accurately. Datasets containing hundreds of thousands of examples can be trained in a few hours, for simpler models. This motivates our interest in advancing the fundamental theory and practice of kernel learning for an efficient class of approximations with randomized feature dimension, with linear scaling in the number of examples, and with predictable convergence rates.

\begin{table*}[htbp]
\begin{center}
\begin{tabular}{|c|c|c|c|}
\hline
\textbf{kernel} & \textbf{parametric form} ($k$) & \textbf{probability distribution} ($\mu$) & \textbf{quantile} ($\gamma$)\\
\hline
$gaussian$ & $\frac{1}{\sigma \sqrt{2\pi}} e^{-\frac{\| x - y \|^{2}}{2\sigma^{2}}}$     & $\frac{\sigma}{\sqrt{2\pi}} e^{-\frac{\sigma^2 \omega}{2}}$ & $\sigma \sqrt{2} \mbox{erf}^{-1}(2u - 1)$\\
$\textit{skewed}-\chi_2$ & $\frac{2 (x + c)^{\sigma} (y + c)^{\sigma}}{(x + c)^{2\sigma} + (y + c)^{2\sigma}}$  & $\frac{1}{2\sigma} \mathrm{sech}\left(\frac{\pi\omega}{2\sigma}\right)$ & $\sigma \frac{2}{\pi} \log \left( \tan \left( \frac{\pi}{2} u \right) \right)$\\
\textit{skewed intersection} & $\min \left\{ \frac{(x + c)^{\sigma}}{(y + c)^{\sigma}}, \frac{(y + c)^{\sigma}}{(x + c)^{\sigma}} \right\}$ & $\frac{1}{\pi} \frac{\sigma}{\sigma^2 +\omega^2}$ & $\sigma \tan \left( \pi (u - \frac{1}{2}) \right)$ \\
\hline
\end{tabular}
\end{center}
\caption{Examples of kernels that can be efficiently optimized within our framework, presented for the unidimensional case. For the multidimensional case, the Fourier methodology decomposes, hence it requires a simple multiplication of parameters corresponding to each dimension. In the above the random variable $u$ is drawn from the uniform distribution.} \label{t:singlekl}
\end{table*}

\subsection{Single Kernel Learning}
\label{sec:skl}
 First we consider learning strategies for the hyper-parameters $\boldsymbol{\sigma}$ of the kernel $k$ based on the Fourier feature methodology. To achieve this goal we need a learning criterion and an algorithm for optimizing the hyper-parameters. Since model training is much faster due to the linear dependence on the training set size, we can perform more complex parameter learning than classic approaches that usually rely on a grid search procedure, like cross-validation. This efficiency of the linear formulation also allows us to scale parameter learning towards numbers of parameters which are unattainable with classical methods. 
Our approach to kernel learning will optimize the hyper-parameters with respect to the error on a held-out validation set, for models obtained on the training set. This has been shown to be a viable procedure to prevent overfitting\cite{Gehler2009}. 

In the sequel we denote $\mathbf{X}$ and $\mathbf{y}$  the matrix of inputs or covariates (row-wise organized) and $\mathbf{y}$ the matrix of targets on the training set, respectively, whereas $\mathbf{U}$ and $\mathbf{v}$ are the validation inputs and targets, respectively. We use $\boldsymbol{\phi}_{\boldsymbol{\Gamma}}$, introduced earlier, as our random Fourier feature map but we will drop $\boldsymbol{\Gamma}$ to simplify notation. We define $\boldsymbol{\phi}(\mathbf{X})$ to be the feature map applied to all the rows of the matrix $\mathbf{X}$.  

We will learn the hyper-parameters of a kernel ridge regression model $\boldsymbol{\beta}\equiv\boldsymbol{\beta}(\boldsymbol{\sigma})$ (other margin-based training costs, such as hinge loss, logistic loss, etc., can be similarly adopted into the framework) 
\begin{equation} \min_{\beta} \frac{1}{2} \left \| \boldsymbol{\phi}(\mathbf{X}) \boldsymbol{\beta} - \mathbf{y} \right \|_{2}^{2} + \lambda \| \boldsymbol{\beta} \|_{2}^{2}. \end{equation} 

For this problem the optimum can be obtained in closed form as \begin{equation} \boldsymbol{\beta} = \left( \boldsymbol{\phi}(\mathbf{X})^{\top} \boldsymbol{\phi}(\mathbf{X}) + \lambda \mathbf{I}_{d} \right)^{-1} \boldsymbol{\phi}(\mathbf{X})^{\top} \mathbf{y}. \end{equation}

To learn the hyper-parameters $\boldsymbol{\sigma}$ we optimize the squared $l_2$ validation error. Given $f = \boldsymbol{\phi}(\mathbf{U})^T \boldsymbol{\beta} - \mathbf{v}$ and $r(\boldsymbol{\sigma})$ a regularizer for the hyper-parameters, \eg $r(\boldsymbol{\sigma}) = \lVert \boldsymbol{\sigma} \rVert_{2}^2$ we optimize
\begin{equation}
\min_{\boldsymbol{\sigma}} \left \| f \right \|_{2}^2 + r(\boldsymbol{\sigma}) 
\end{equation} 

Note that we can easily obtain a gradient with regard to the kernel parameters for this optimization. The random feature map $\boldsymbol{\phi}$ establishes a connection between the original input representation $\mathbf{X}$ and the approximation of its lifting $\boldsymbol{\phi}(\mathbf{X})$. This can be used to derive analytical expressions not only for kernel expansions in the input space but also for the gradient with regard to the kernel parameters. We can minimize the loss using a local-descent based optimization. 

Given $\sigma_{i}$ the $i$'th dimension of the kernel parameter vector $\boldsymbol{\sigma}$, then 
\begin{eqnarray}
	\frac{\partial \left \| f \right \|_{2}^2}{\partial \sigma_{i}} &=&
			\mbox{Tr}\left[ \left( \frac{\partial \left \| f \right \|_{2}^{2} }{\partial f}\right)^{\top} \frac{\partial f}{\partial \sigma_{i}}\right] \\
&=& f^{\top} \frac{\partial f}{\partial \sigma_{i}} =f^{\top} \left(\frac{\partial \boldsymbol{\phi}(\mathbf{U})}{\partial \sigma_{i}} \boldsymbol{\beta} + \boldsymbol{\phi}(\mathbf{U}) \frac{\partial \boldsymbol{\beta}}{\partial \sigma_{i}}\right) \nonumber
\end{eqnarray}

For translation invariant kernels, we can easily compute $\boldsymbol{\phi}_{\Gamma}$. Manipulating the sampling distribution $\mu_k$ associated with kernel $k$ in an optimization process identifies the different frequency bands that are useful for prediction on training data. This effectively leads to a posterior frequency distribution, given $\mu_k$ as a prior. Generating samples from this posterior and optimizing with respect to their expectation on the training set gives a direct way of learning the parameters of the kernel. However as the kernel hyper-parameters $\boldsymbol{\sigma}$ change, $\boldsymbol{\Gamma}$ needs to be re-sampled. Although this is feasible, changing the sampling distribution introduces conceptual difficulties during optimization as the objective function change, and sampling becomes a source of noise and non-smoothness for the optimization. Importance sampling can be used to avoid resampling at each step, but if the parametrized frequency distribution drifts far away from the starting distribution, then the original samples will have little importance weights, and resampling would potentially be needed nevertheless for convergence. Such difficulties can be overcome for several interesting kernels that belong to a class of functions where the samples from $\mu_k$ can be written as \begin{equation}\boldsymbol{\gamma} = \boldsymbol{\sigma} \cdot h(\boldsymbol{\omega})\label{gamma_form}\end{equation} where $h$ is the quantile function, $\boldsymbol{\omega}$ are uniformly sampled and fixed, and $\cdot$ is the Hadamard product of two vectors. In this case, throughout the entire optimization process, sampling needs to be done only once from the uniform distribution for $\omega$. When $\boldsymbol{\sigma}$ is changed, samples from the new distribution are generated automatically from $\boldsymbol{\omega}$.
Examples of such kernels are the \textit{Gaussian}, the  \textit{generalized skewed-$\chi_2$} and the \textit{generalized skewed intersection}\cite{li_dagm10} (see table \ref{t:singlekl}).
	
Based on this property, by differentiating the feature \mbox{map  \eqref{general_feat_formula}} we obtain

\begin{eqnarray}
 \frac{\partial \boldsymbol{\phi}(\mathbf{u}_{k})}{\partial \sigma_{i}} = \sqrt{\frac{2}{d}} \left[ \frac{\partial \cos(\mathbf{u}_{k}^{\top} \left( \boldsymbol{\sigma} \cdot h(\boldsymbol{\omega}_{j})\right) + 2\pi b_{j})}{\partial \sigma_{i}}\right]_{j = 1, d} && \label{eq:DPhiofUatDomega} \\
= \sqrt{\frac{2}{d}} \left[ -\mathbf{u}_{k,i}^{\top} h(\boldsymbol{\omega}_{j,i}) \sin(\mathbf{u}_{k}^{\top} \left( \boldsymbol{\sigma} \cdot h(\boldsymbol{\omega}_{j})\right) + 2\pi b_{j})\right]_{j = 1, d} && \nonumber
\end{eqnarray}

To compute the second term $\frac{\partial \boldsymbol{\beta}}{\partial \sigma_{i}}$, we first define the matrix $\mathbf{Q} = \phi(\mathbf{X})^{\top} \phi(\mathbf{X}) + \lambda \mathbf{I}_{d}$ and using the standard result $\frac{\partial \mathbf{Q}^{-1}}{\partial \sigma_{i}} = -\mathbf{Q}^{-1} \frac{\partial \mathbf{Q}}{\partial \sigma_{i}} \mathbf{Q}^{-1}$, we obtain 
\begin{eqnarray}
\frac{\partial \boldsymbol{\beta}}{\partial \sigma_{i}}
&=&\frac{\partial \mathbf{Q}^{-1}}{\partial \sigma_{i}} \boldsymbol{\phi}(\mathbf{X})^{\top} \mathbf{y} + \mathbf{Q}^{-1} \left(\frac{\partial \boldsymbol{\phi}(\mathbf{X})}{\partial \sigma_{i}}\right)^{\top}\mathbf{y} \label{eq:DbetaDomega} \\
&=&  -\mathbf{Q}^{-1} \frac{\partial \mathbf{Q}}{\partial \sigma_{i}} \mathbf{Q}^{-1} \boldsymbol{\phi}(\mathbf{X})^{\top} \mathbf{y} + \mathbf{Q}^{-1} \left(\frac{\partial \boldsymbol{\phi}(\mathbf{X})}{\partial \sigma_{i}}\right)^{\top}\mathbf{y} \nonumber
\end{eqnarray}
It is easy to see that $\frac{\partial \phi(\mathbf{X})}{\partial \sigma_{i}}$ can be computed in the same way as in equation \eqref{eq:DPhiofUatDomega}, as the gradient of $\mathbf{Q}$ with respect to $\sigma_{i}$ can be obtained as
\begin{eqnarray}
	\frac{\partial \mathbf{Q}}{\partial \sigma_{i}} = \left(\frac{\partial \boldsymbol{\phi}(\mathbf{X})}{\partial \sigma_{i}}\right)^{\top} \boldsymbol{\phi}(\mathbf{X}) + \boldsymbol{\phi}(\mathbf{X})^{\top} \frac{\partial \boldsymbol{\phi}(\mathbf{X})}{\partial \sigma_{i}}. 
\end{eqnarray}
Computing the gradient of $r(\boldsymbol{\sigma})$ will depend on the type of regularization chosen. In general this is a smooth function of $\boldsymbol{\sigma}$ (\eg $l_{2}$ norm) so it will be straightforward to compute $\frac{\partial r(\boldsymbol{\sigma})}{\partial \sigma_{i}}$. 

	Now that we have a closed form formula for the gradient with respect to all kernel parameters in the Fourier domain we can plug it into a non-linear optimizer and estimate the parameters. 

	This overall philosophy bears a certain similarity to the objective introduced in \cite{Chapelle2002} where the authors use a gradient descent learning technique for kernel ridge regression based on the exact kernel matrix. It is also similiar to the multiple kernel learning technique for products of kernels introduced in \cite{Varma09}.  However, besides the technical differences that are introduced by the usage of a Fourier embedding map, an important advantage for our methodology is that we do not have to store the kernel matrices in memory which has $O(N^2)$ memory cost. Our memory requirement is just $O(Nd+d^2)$ where $d$ is the size of the Fourier embedding. 
The computational complexity of our method is dominated by $O(i_{skl}(Nd^2+d^3+rNd))$ where $d$ is the size of the random Fourier features, $N$ is the number of training samples, $r$ is the number of parameters and $i_{skl}$ is the number of iterations to convergence. $O(Nd^2 + d^3)$ is the cost of computing matrix $\mathbf{Q}$ and inverting it.

\subsection{Multiple Kernel Learning}
\label{sect:mkl}
	Previous work has proven an underlying connection between multiple kernel learning and group Lasso\cite{consistencyBach}.  Although this is an interesting property, it has found relatively limited practical applications so far. In this section we show that by using the random Fourier methodology, we can do just the opposite and lift group Lasso to the kernel domain. We propose a multiple kernel learning formulation where the features are initially transformed using the random Fourier framework, concatenated, and group Lasso is applied to them (RFF-GL). We prove that this new formulation is equivalent with the multiple kernel learning formulation introduced in \cite{Varma09} and then compare the two methodologies. Experiments show that both approaches have similar performance. In contrast, group Lasso based on random Fourier features runs faster and scales better since we do not need to compute or store the Gramm matrices associated with the features. 

Let $\mathbf{X}$ be the input matrix for the training set organized row-wise, $\mathbf{y}$ the associated targets and $\{k_{1}, \ldots, k_{r}\}$ a set of kernels. For example $\mathbf{X}$ could be a concatenation of multiple image features such as SIFT or HOG and therefore should be represented as $\left\{ \mathbf{X}_{1}, \ldots, \mathbf{X}_{r} \right\}$. But to simplify the notation we would not refer to the individial components and will just use $\mathbf{X}$ when we refer to the input. We denote $\mathbf{\mathcal{F}}_{i}$ the matrix of random Fourier features obtained from approximating $k_{i}$ on inputs $\mathbf{X}$. We concatenate $\mathcal{F}_i$ column-wise to obtain a matrix $\mathcal{F}$ on which we can apply group Lasso with $r$ non-overlapping groups, each corresponding to the Fourier embedding of a different kernel. We use the $l_1$-$l_2$ formulation \cite{Yuan06modelselection} which can be written as:
\begin{equation}\min_{\mathbf{w}} \lambda  \displaystyle \sum_{i = 1}^{r} \displaystyle \| \mathbf{w}_{\mathcal{F}_{i}} \|_{2} + l(\mathbf{y}, \mathcal{F}\mathbf{w}), \end{equation}
where $\mathbf{w}_{\mathcal{F}_{i}}$ are the weights applied to the features $\mathcal{F}_{i}$ and $l(y, f(x))$ is a loss function, which can be a quadratic loss or an approximation for the $\epsilon$-insensitive regression loss. In our  experiments the optimization was performed with a group Lasso solver \cite{Liu:2009:SLEP:manual} which was adapted for our specific loss functions.

	We compare the above formulation with the standard multiple kernel learning method (GMKL)  presented in \cite{Varma09} where we consider the regression problem under $l_{1}$ regularization. For GMKL we have to build the Gram matrices $\{\mathbf{K}_{i} \}_{i = 1..r}$ for each kernel $k_{i}$ on $\mathbf{X}$. We focus on the regression problem and define 
$\mathbf{K}_{\mathbf{d}} = \displaystyle \sum_{ i = 1..r} \mathbf{d}_i \mathbf{K}_{i}$. The algorithm will output the optimized values for $\mathbf{d}$ and a set of support vectors and both these elements will be leveraged in the model used for testing.

\subsubsection{Proof of equivalence}	
	
	We now show that GMKL is equivalent as an optimization procedure with RFF-GL. For GMKL we have the following primal problem
\begin{eqnarray}
\min_{\mathbf{w}, \mathbf{d}}\frac{1}{2}\mathbf{w}^{\top}\mathbf{w} &+& C\sum_{i=1}^{n} l(y_{i}, \psi(\mathbf{x}_{i})^{\top} \mathbf{w}) + \sum_{t=1}^{r} d_{t} \label{enqEqv} \nonumber \\
 \mbox{subject to} && \mathbf{d} \geq 0
\end{eqnarray}
where 
\begin{eqnarray}
 \psi(\mathbf{x}_{i}) = [ \sqrt{d_{1}} \psi_{1}(\mathbf{x}_{i})^{\top}, \ldots, \sqrt{d_{r}} \psi_{r}(\mathbf{x}_{i})^{\top}]^{\top}
\end{eqnarray}

and we defined $\psi_{t}(\mathbf{x})$ as the feature embedding associated to kernel $k_{t}$. 

	From the Representer Theorem \cite{Scholkopf2002} we now show that the solution for $\mathbf{w}$ will be a linear combination of the basis functions 
\begin{eqnarray}
	\mathbf{w} &=&  \sum_{i = 1}^{n} \alpha_{i} \psi(\mathbf{x}_{i}) \\
	&=& \left[ \sqrt{d_{1}}\mathbf{w}_{1}^{\top}, \ldots, \sqrt{d_{k}}\mathbf{w}_{k}^{\top} \right]^{\top} \nonumber
\end{eqnarray}
where we have dropped the bias term for simplicity. We can rewrite the above optimization problem
\begin{eqnarray}
&& \hspace {-.75cm} \min_{\mathbf{w}, \mathbf{d}} \sum_{t = 1}^{k} \frac{d_{t}}{2} \lVert \mathbf{w}_{t} \rVert_{2}^{2} + C\sum_{i=1}^{n} l(y_{i}, \sum_{t = 1}^{k} d_{t} \psi_{t}(\mathbf{x}_{i})^{\top} \mathbf{w}_{t}) + \sum_{t=1}^{k} d_{t} \nonumber \\
&& \mbox{subject to } \mathbf{d} \geq 0
\end{eqnarray}

Following a standard trick from the multiple kernel learning literature\cite{consistencyBach} we make the substitutions $\mathbf{w}_{t} \rightarrow d_{t}\mathbf{w}_{t}$ to obtain

\begin{eqnarray}
&&  \min_{\mathbf{w}, \mathbf{d}}\sum_{t = 1}^{k} \frac{\lVert \mathbf{w}_{t} \rVert_{2}^{2}}{2d_{l}} + C\sum_{i=1}^{n} l(y_{i}, \sum_{t = 1}^{k} \psi_{t}(\mathbf{x}_{i})^{\top} \mathbf{w}_{t}) + \sum_{t=1}^{k} d_{t} \nonumber \\
&&  \mbox{subject to } \mathbf{d} \geq 0
\end{eqnarray}

It can easily be shown that 
\begin{equation}
\frac{1}{2} \frac{\lVert \mathbf{w}_{t} \rVert_{2}^{2}}{ d_{t}} + d_{t} \geq \sqrt{2} \lVert \mathbf{w}_{t} \rVert_{2} \label{eq:etaTrick}
\end{equation}
and we observe that the loss function no longer depends on $\mathbf{d}$. Therefore the primal multiple kernel learning formulation is equivalent to the following group Lasso formulation

\begin{eqnarray}
\min_{\mathbf{w}} \lambda \sum_{t = 1}^{k} \lVert \mathbf{w}_{t} \rVert_{2} + \sum_{i=1}^{n} l(y_{i}, \sum_{t = 1}^{k} \psi_{t}(\mathbf{x}_{i})^{\top} \mathbf{w}_{l})
\end{eqnarray}
where we set $\lambda = \frac{\sqrt{2}}{C}$. The equality happens when $d_{t} = \frac{1}{\sqrt{2}} \lVert \mathbf{w}_{t} \rVert_{2}$. This approach is suitable for other types of regularization for the parameters $\mathbf{d}$ as well, not only the $l_{1}$ norm which we have used. The only restriction is that we should be able to find a closed form solution for \eqref{eq:etaTrick}.

Now we consider the loss function which has to be differentiable in order to be suitable for a group Lasso formulation. The $\epsilon$-insensitive loss ($\epsilon$-IL) used in standard support vector regression is not differentiable but we can use a smooth approximation instead. In our case we have selected an $\epsilon$-insensitive $\gamma$ logistic loss ($\epsilon$-IGLL) function defined as \cite{Dekel03smoothepsilon-insensitive, RennieF04}
\begin{eqnarray}
l_{\gamma, \epsilon} (y_{i},f(\mathbf{x})) &=& \frac{1}{\gamma} \log\left(1 + e^{\gamma\left(f(\mathbf{x}) - y_{i} - \epsilon\right)}\right) \\ && 
+ \frac{1}{\gamma} \log\left(1 + e^{\gamma\left(-f(\mathbf{x}) + y_{i} - \epsilon)\right)}\right) \nonumber \\ &&
- \frac{2}{\gamma} \log(1 + e^{-\gamma \epsilon}) \nonumber
\end{eqnarray}

We could also consider quadratic or logistic losses for solving the group Lasso.
\begin{figure}[h]
\begin{center}
  \scalebox{0.35}{\includegraphics[viewport=1.5cm 6.75cm 19.25cm 20.75cm]{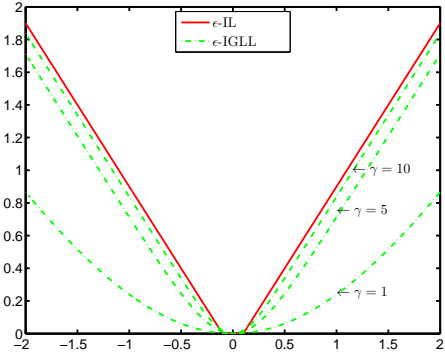}}
\end{center}
   \caption{The $\epsilon$-insensitive $\gamma$ loss function we use for optimization provides a good approximation to the exact $\epsilon$-insensitive loss. Here we show values for $\epsilon=0.1$ and $\gamma = 1,5,10$}
\label{fig:eil}
\end{figure}

\subsubsection{Computational complexity}
	
	We assume an average time complexity for a support vector machine algorithm \cite{CC01a} to be $O(N_{sv}^{3})$ where $N_{sv}$ is the number of support vectors. Since GMKL is based on several evaluations of the standard SVM solver we can show the time complexity of the multiple kernel learning framework is $O \left( rN^2m^2 + i_{GMKL}(N^2r+N_{sv}^{3}) + rN_{sv}Nm^2\right)$ where $r$ is the number of kernels, $N$ is the number of training samples, $m$ is the size of the training input (we assume it is the same for all data for simplicity) and $i_{GMKL}$ is the maximum number of calls to and SVM solver. The complexity is dominated by the term $O(rN^2m^2)$ which is the cost of computing the kernel matrices.

	On the other hand for the group Lasso formulation we have complexity $O(rNm^2d + i_{gl}Ndr)$ where $d$ is the size of the random Fourier features and $i_{gl}$ is the number of iterations required for group Lasso. If we use an approximated kernel as in \cite{sreekanth10generalized} another $(2p+1)^2$ cost is added to the preprocessing step. The complexity  becomes $O(rNm^2(2p+1)^2d)$ where $2p+1$ is the dimension of the approximated kernel as discussed in \cite{Vedaldi2010}. The memory complexity is $O(Nrd)$ which is definitely smaller than $O(rN^2)$ for GMKL. 
	
	We see that our algorithm scales linearly, whilst the nonlinear kernel method is quadratic. The constants matter because for a small number of samples our method could run slower given that $d$ is in the range of thousands, $i_{GMKL}$ is usually $10^2$ and $i_{gl}$ is in the order of $d$.
	 
\section{Experiments}

	We present experiments for single kernel learning and multiple kernel learning, comparing the random linear Fourier methodology with its non-linear kernel counterparts both in terms of running times and in terms of accuracy. 

	We have experimented with the PASCAL VOC2011  segmentation challenge. This consists of 2223 images for training with ground truth split into halves for training and validation. Another 1111 images are given for testing without ground truth. Following the standard procedure we have trained the methods on the training set (further split, internally into training and validation for kernel hyper-parameter learning) and tested on the PASCAL VOC2011 validation set. We have used the methodology presented in Li \etal \cite{lis_cvpr10} and relied on a segmentation algorithm from Carreira and Sminchisescu \cite{carreira_cvpr10} to filter the initial pool of segments to around $100$ segments per image. For each of these segments we extracted 8 types of features among which two bag of visual words for color SIFT \cite{vandeSandeTPAMI2010} and two dense gray scale SIFT descriptors one on each segment and one on the background of each segment, three types of phog descriptors two on the pb edges given by \cite{globalPB} computed at different scales, one on the contour and one on the foreground. The third phog descriptor uses no pb. The last descriptor is a pyramid of locally binary pattern features \cite{Ojala02multiresolutiongray-scale} for texture classification. Because the number of segments (around $10^{5}$) was still too large for any kernel support vector based algorithm to cope with, we have chosen up to $10^4$ segments for each class. The segments were chosen based on their individual scores and we balanced the examples to have a fair split between positive and negative segments. More details on how the scores are defined and computed could be found in \cite{lis_cvpr10}.

\subsection{Single kernel learning}

	We ran a set of experiments for our random Fourier features single kernel learning technique (RFF-SKL) and compared in terms of accuracy with the single kernel learning technique introduced by Chapelle \etal \cite{Chapelle2002} (KRR-GD). We want to predict the class for more than $10^5$ segments. We expect from RFF-SKL to give comparable results in terms of accuracy to KRR-GD. Results are shown in Table \ref{t:skl-time}. 

\begin{figure}[h]
\begin{center}
  \scalebox{0.35}{\includegraphics[viewport=1.5cm 6.75cm 19.25cm 20.75cm]{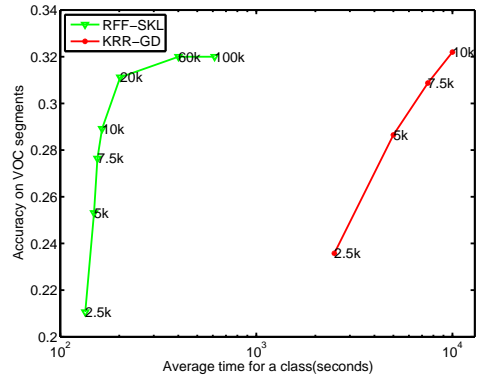}}
\end{center}
   \caption{Time versus accuracy for classifying the VOC segments for RFF-SKL and KRR-GD.  We varied the number of training points from $10^3$ up to $10^5$ and measured the average time for a class, for 20 classes. For KRR-GD it was not feasible to go beyond $10^4$ training samples. x axis is log scale.}
\label{fig:voc-acc-skl}
\end{figure}

We vary the size of the training set from $10^3$ to $10^5$ and measured both the running times of RFF-SKL and the accuracy. In Figure \ref{fig:voc-acc-skl} we present how the accuracy depends on the size of the training data and the average time required for  RFF-SKL and KRR-GD to run on a class. We observe that RFF-SKL running time scales linearly in the number of examples. The number of random Fourier samples we have chosen was $d=3000$. This is consistent with our computational complexity discussed in Section \ref{sec:skl}. We are able to tune the hyper-parameters of a model trained with more than $10^5$  samples, each with 3000 attributes in less than 15 minutes.

\begin{table}[h]
\begin{center}
\scalebox{0.8} {
\begin{tabular}{|l|c|c|c|c|c|c|c|}
\hline
Samples & $2.5\cdot 10^3$ & $5\cdot 10^3$ & $7.5\cdot 10^4$ & $10^4$ & $2\cdot 10^4$ & $6\cdot 10^4$\\
\hline
RFF-SKL & $0.21$ & $0.25$ & $0.28$ & $0.29$ & $0.31$ & $0.32$\\
\hline
KRR-GD & $0.24$ & $0.29$ & $0.31$ & $0.32$ & * & * \\
\hline
\end{tabular}
}
\end{center}
\caption{Accuracy of RFF-SKL versus KRR-GD. For a large number of training samples the nonlinear method could not be run due to memory limits.}
\label{t:skl-time}
\end{table}

\subsection{Multiple Kernel Learning} 

\begin{figure}[h]
\begin{center}
  \scalebox{0.35}{\includegraphics[viewport=1.5cm 6.75cm 19.25cm 20.75cm]{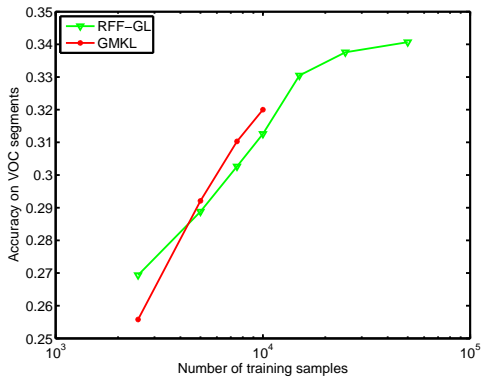}}
\end{center}
   \caption{Accuracy for classifying the VOC segments for RFF-GL and GMKL, as a function of the number of training examples. RFF-GL scales significantly better than GMKL. }
\label{fig:voc-acc-mkl}
\end{figure}

	We also report experiments for multiple kernel learning. For GMKL we have evaluated an exponentiated $\chi^2$  kernel for each image feature. We set the scaling parameter to be the mean of the chi-square distance matrix following the procedure from \cite{Gehler2009}. The Gramm matrices were created for each class since for different classes we had to select different representative samples. For our random Fourier features within the group Lasso framework (RFF-GL) we have approximated the kernel as in \cite{sreekanth10generalized} using the recommended settings. In Figure \ref{fig:voc-acc-mkl} we compare the accuracy of predicting the right class for the segments. We see that for a small number of training samples GMKL is slightly superior, but RFF-GL catches up due to its scalability. Working with kernel matrices larger than $10^4$ was not feasible for GMKL. 

For group Lasso we have adapted the implementation presented in \cite{Liu:2009:SLEP:manual} whereas for comparisons with standard multiple kernel learning, we have used the GMKL implementation presented by Varma and Babu \cite{Varma09}. 

\begin{figure}[h]
\begin{center}
  \scalebox{0.35}{\includegraphics[viewport=1.5cm 6.75cm 19.25cm 20.75cm]{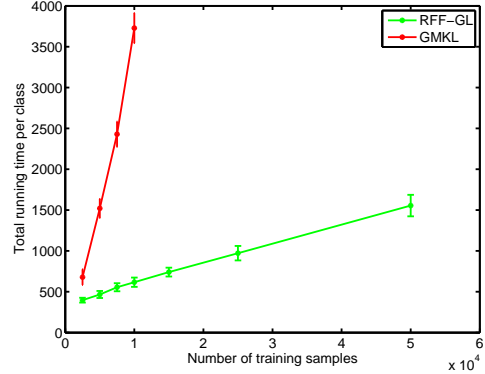}}
\end{center}
   \caption{Running times expressed in seconds for RFF-GL and GMKL. Notice that RFF-GL scales linearly whereas GMKL scales quadratically. Error-bars are obtained by averaging over the number of classes (20 in this case).}
\label{fig:voc-time-mkl}
\end{figure}

In Figure \ref{fig:voc-time-mkl} we show the average running times for a class. We see that RFF-GL scales linearly and GMKL scales quadratically. 

Following the relation given by eq. \eqref{eq:etaTrick} in Table \ref{t:coefficients} we compare the weights given by the two methods on one of the classes on which we have performed regression -- in this case the aeroplane class. We see that both RFF-GL and GMKL favour the SIFT over pHOG.

\begin{table*}[htb]
\begin{center}
\begin{tabular}{|c|c|c|c|c|c|c|c|c|}
\hline
& pHOG 1 & pHOG 2 & pHOG no pb & SIFT fg & SIFT bgnd & color SIFT fg & color SIFT bgnd & LBP \\
\hline
RFF-GL & 0.2398 & 0.2260 & 0.5652 & 1.7903 & 1.0495 & 1.1287 & 0.8505 & 0.5847 \\
GMKL & 0.3862 & 0.1901 & 0.6428 & 1.5679 & 0.5333 & 0.4243 & 0.3319 & 0.1081 \\
\hline
\end{tabular}
\end{center}
\caption{For the RFF-GL we show the $l_{2}$ norm of the coefficients corresponding to each group, whereas for GMKL we show the coefficients $\mathbf{d}$ mentioned in section \ref{sect:mkl}. These are for the aeroplane class from VOC. We see both methods tend to associate similar weights to the different image features, and also the quantitative relationship given by eq. \eqref{eq:etaTrick} is preserved.}  \label{t:coefficients}
\end{table*}

\section{Conclusions}

Fourier methodology is a powerful and formally consistent class of linear approximation techniques for non-linear kernel machines that carries the promise of combining good model scalability and non-linear prediction power. This has motivated research in extending the class of useful kernels that can be approximated, e.g. Chi-square\cite{Vedaldi2010,li_dagm10}, but leaves ample space for reformulating standard problems like single or multiple kernel learning in the linear Fourier domain. In this paper we have developed gradient-based methods for single and multiple-kernel learning in the Fourier domain and showed that these are efficient and produce accurate results on a complex computer vision dataset like VOC2011. In future work we plan to explore alternative kernel basis expansions, feature selection and non-convex optimization techniques for learning.

{\small
\bibliography{Bibliography}
}
\end{document}